\documentclass[twocolumn]{bmcart}

\usepackage{amsthm,amsmath,bm}
\usepackage[colorlinks,linkcolor=blue]{hyperref}
\usepackage[utf8]{inputenc} 
\usepackage{multirow,makecell}
\usepackage{enumerate}
\usepackage{tabularx}

\usepackage[pdftex]{graphicx}

 \usepackage{array}

\usepackage{dblfloatfix}

\startlocaldefs
\endlocaldefs

\begin{document}

\begin{frontmatter}

\begin{fmbox}
\dochead{Research}


\title{No-Reference Color Image Quality Assessment: From Entropy to Perceptual Quality}


\author[
   addressref={aff1},                   
   email={cxq@whu.edu.cn}   
]{\inits{XC}\fnm{Xiaoqiao} \snm{Chen}}
\author[
   addressref={aff1},
   email={zhqy@whu.edu.cn}
]{\inits{QZ}\fnm{Qingyi} \snm{Zhang}}
\author[
   addressref={aff1},
   email={linmanhui@whu.edu.cn}
]{\inits{ML}\fnm{Manhui} \snm{Lin}}
\author[
   addressref={aff1},
   corref={aff1},
   email={ygy@whu.edu.cn}
]{\inits{GY}\fnm{Guangyi} \snm{Yang}}
\author[
   addressref={aff1},
   email={chuhe@whu.edu.cn}
]{\inits{CH}\fnm{Chu} \snm{He}}


\address[id=aff1]{
  \orgname{School of Electronic Information, Wuhan University}, 
  \postcode{430072}                                
  \city{Wuhan},                              
  \cny{China}                                    
}
\address[id=aff2]{%
  \orgname{Collaborative Innovation Center of Geospatial Technology, Wuhan University},
  \postcode{430079}
  \city{Wuhan},
  \cny{China}
}





\begin{abstractbox}

\begin{abstract} 
This paper presents a high-performance general-purpose no-reference (NR) image quality assessment (IQA) method based on image entropy. The image features are extracted from two domains. In the spatial domain, the mutual information between the color channels and the two-dimensional entropy are calculated. In the frequency domain, the two-dimensional entropy and the mutual information of the filtered sub-band images are computed as the feature set of the input color image. Then, with all the extracted features, the support vector classifier (SVC) for distortion classification and support vector regression (SVR) are utilized for the quality prediction, to obtain the final quality assessment score. The proposed method, which we call entropy-based no-reference image quality assessment (ENIQA), can assess the quality of different categories of distorted images, and has a low complexity. The proposed ENIQA method was assessed on the LIVE and TID2013 databases and showed a superior performance. The experimental results confirmed that the proposed ENIQA method has a high consistency of objective and subjective assessment on color images, which indicates the good overall performance and generalization ability of ENIQA. The source code is available on github~\href{https://github.com/jacob6/ENIQA}{https://github.com/jacob6/ENIQA}.
\end{abstract}


\begin{keyword}
\kwd{Image entropy}
\kwd{Mutual information}
\kwd{No-reference image quality assessment}
\kwd{Support vector classifier}
\kwd{Support vector regression}
\end{keyword}


\end{abstractbox}
\end{fmbox}

\end{frontmatter}



\section{Introduction}  \label{sect:intro}
In this era of information explosion, we are surrounded by an overwhelming amount of information. The diversification of information is dazzling, and images, as the source of visual information, contain a wealth of valuable information. Considering the incomparable advantages of image information over other types of information, it is important to process images appropriately in the different fields~\cite{mohammadi2014subjective}. In image acquisition, processing, transmitting, and recording, image distortion and quality degradation are an inevitable result of the imperfection of the imaging system, the processing method, the transmission medium, and the recording equipment, as well as object movement and noise pollution~\cite{fang2014objective,zhang2017realizing,gu2018learning}. Image quality has a direct effect on people's subjective feelings and information acquisition. For example, the quality of the collected images directly affects the accuracy and reliability of the recognition results in an image recognition process~\cite{fronthaler2006automatic}. Another example is that remote conferencing and video-on-demand systems are affected by such factors as transmission errors, network latency, and so on~\cite{li2009reduced}. Online real-time image quality control is thus introduced to ensure that the service provider dynamically adjusts the source location strategy, so as to meet the service quality requirements~\cite{hemami2010no}. It is therefore not surprising that research into image quality assessment (IQA) has received extensive attention during the last two decades~\cite{chandler2013seven}.

In accordance with the need for human participation, IQA methods can be divided into two classes: subjective image quality assessment methods and objective image quality assessment methods~\cite{wang2004image}. Subjective assessment is quantified by the human eye. In contrast, an objective IQA method focuses on automatic assessment of the images via a specific method by the use of computing equipment, with the ultimate goal of enabling a computer to act as a substitute for the human visual system (HVS) in viewing and perceiving images~\cite{lin2011perceptual}. In practice, subjective assessment results are difficult to apply in real-time imaging systems due to their strong randomicity. Therefore, objective IQA methods have been widely studied~\cite{moorthy2011visual}. According to the availability of a reference image, objective IQA methods can be classified as full-reference (FR), reduced-reference (RR), and no-reference (NR) methods~\cite{bovik2013automatic}. In an FR method, an original “distortion-free” image is assumed to be supplied, as the assessment result is obtained through the comparison of the two images. With the advances of recent studies, the accuracy of this kind of method is getting better, despite its disadvantage of requiring a complete reference image, which is often not available in practical applications~\cite{xue2014gradient}. An RR method, which is also known as a partial reference method, does not make a complete comparison between the distorted image and the pristine one, but only compares certain features~\cite{wang2011reduced}. Conversely, an NR method, which is also called a blind image quality assessment (BIQA) method, requires no image as reference. Instead, the quality is estimated according to the features of the distorted image~\cite{bovik2013automatic}. In many practical applications, a reference image will be inaccessible, and thus the NR-IQA methods have the most practical value and a very wide application potential~\cite{xu2017no-reference/blind}.

In general, the current NR-IQA methods can be divided into two categories: application-specific and general-purpose assessment~\cite{kamble2015no}. The former kind of method assesses the image quality of a specific distortion type and calculates the corresponding score. Common types of distortion include JPEG, JPEG2000 (JP2K), blur, contrast distortion, and noise. For images with compression degradation, Suthaharan $et~al.$~\cite{suthaharan2009no} proposed the visually significant blocking artifact metric (VSBAM) to estimate the degradation level caused by compression. For images with blur degradation, Ciancio $et~al.$~\cite{ciancio2011no} utilized various spatial features and adopted a neural network model to assess the quality. The maximum local variation (MLV) method proposed by Khosro $et~al.$~\cite{bahrami2014fast} provides a fast method of blur level estimation. Rony $et~al.$~\cite{ferzli2009no} put forward the concept of just noticeable blur (JNB) and the improved version of cumulative probability of blur detection (CPBD)~\cite{narvekar2011no}. For images with contrast distortion, Fang $et~al.$~\cite{fang2015no} extracted features from the statistical characteristics of the 1-D image entropy distribution and developed an assessment model based on natural scene statistics (NSS)~\cite{ruderman1994statistics}. Hossein $et~al.$~\cite{nafchi2018efficient} used higher orders of the Minkowski distance and entropy to apply an accurate measurement of the contrast distortion level. For images with noise, Yang $et~al.$~\cite{yang2017no} proposed frequency mapping (FM) and introduced it into quality assessment. Gu $et~al.$~\cite{gu2017no} proposed a training-free blind quality method based on the concept of information maximization. These methods, however, require prior knowledge of the distortion type, which limits their application range. Therefore, general-purpose NR-IQA methods based on training and learning are highly desirable.

General-purpose NR-IQA methods can be further divided into two types: explicit methods and implicit methods~\cite{kim2017deep}. An explicit method usually contains two steps: feature extraction and model mapping~\cite{guan2017visual}. Generally speaking, the features extracted in the first step represent the visual quality, while the mapping model in the second step bridges the gap between the features and the ground-truth quality score. An implicit general-purpose NR-IQA method constructs a mapping model via deep learning. Although deep networks nowadays generally have an independent feature extraction capability, it is difficult for the existing IQA databases to meet the huge demand for training samples, let alone the large amount of redundant data and network parameters. In addition, compared to preselected features, no clear physical meaning can be given by these automatically extracted features. Thus, manual feature extraction is still an effective and accurate way to summarize the whole image distortion.

According to the existing literature, the features extracted by explicit general-purpose NR-IQA methods are mainly concentrated in two categories. $1)$ The parameters of a certain model are obtained after a preprocessing operation such as mean-subtracted contrast-normalized (MSCN) coefficients~\cite{gu2015using}. The typical models are the generalized Gaussian distribution (GGD) model~\cite{moorthy2010two}, the asymmetric GGD (AGGD) model~\cite{mittal2012no}, the Weibull distribution (WD) model~\cite{zhang2015feature}, etc. $2)$ Physical quantities that reflect the characteristics of the image are obtained after preprocessing such as blocking and transformation. The typical methods are image entropy~\cite{liu2014no}, wavelet sub-band correlation coefficients~\cite{moorthy2011blind}, etc. The mapping models from features to image quality are divided into three main types. $1)$ Classical methods such as BIQI~\cite{moorthy2010two}, DIIVINE~\cite{moorthy2011blind}, DESIQUE~\cite{zhang2013no}, and SSEQ~\cite{liu2014no} follow a two-stage framework. The probability of each type of distortion in the image is gauged by a support vector classifier (SVC) and denoted as $p_i$ in the first stage. The quality of the image along each of these distortions is then assessed by support vector regression (SVR) and denoted as $q_i$ in the second stage. Finally, the quality of the image is expressed as a probability-weighted summation: $Index=\sum p_iq_i$. $2)$ Methods such as NIQE~\cite{mittal2013making} and IL-NIQE~\cite{zhang2015feature} are classified as “distortion-unaware”, and they calculate the distance between a model fitted by features from a distorted image and an ideal model to estimate a final quality score, without identifying the type of distortion. $3)$ Methods such as BLIINDS-II~\cite{saad2012blind}, and BRISQUE~\cite{mittal2012no} implement direct mapping of the image features to obtain a subjective quality score, also without distinguishing the different distortion types.

The existing general-purpose NR-IQA methods are faced with the following problems. $1)$ The color space of the image is less considered in these methods. $2)$ Some of the methods take advantage of only the statistical features of the pixels, and they ignore the spatial distribution of the features. Liu $et~al.$~\cite{liu2014no} calculated the 1-D entropy of image blocks in the spatial and frequency domains, respectively, and used the mean, along with the skewness~\cite{motoyoshi2007image}, of all the local entropy values as the image features to implement the SSEQ method. Gabarda $et~al.$~\cite{gabarda2007blind} approximated the probability density function by the spatial and frequency distribution to calculate the pixel-wise entropy on a local basis. The measured variance of the entropy is a function of orientation, which is used as an anisotropic indicator to estimate the fidelity and quality of the image~\cite{dong2015exploiting}. Although some aggregated features of image grayscale distribution can be embodied in these one-dimensional entropy-based methods, the spatial features of the distribution cannot be obtained.

In this paper, we introduce an NR-IQA method based on image entropy, namely, ENIQA. Firstly, by using the two-dimensional entropy (TE)~\cite{abutaleb1989automatic} instead of the one-dimensional entropy~\cite{brink1996using}, the proposed method better embodies the correlativity of pixel neighbors. Secondly, we calculate the mutual information (MI)~\cite{maes1997multimodality} between the different color channels and the TE of the color image in two scales. we split the image into patches in order to exploit the statistical laws of each local region. During this process, visual saliency detection~\cite{zhang2016application} is performed to weight the patches, and the less important ones are then excluded. Thirdly, a Log-Gabor filter~\cite{field1987relations,zhang2011fsim} is applied on the image to simulate the neurons' selective response to stimulus orientation and frequency. After that, the MI between the different sub-band images and the TE of the filtered images are computed. The MI, as well as the mean and the skewness of the TE, is then utilized as the structural feature to determine the perceptual quality of the input image. Specifically, SVC and SVR are used to implement a two-stage framework for the final prediction. The experiments undertaken with the LIVE~\cite{sheikh2005live} and TID2013~\cite{ponomarenko2015image} databases confirmed that the proposed ENIQA method performs well and shows a high consistency of subjective and objective assessment.

The rest of this paper is structured as follows. In Section~\ref{sec:proposed}, we introduce the structural block diagram of the novel IQA method proposed in this study, and we then present a detailed introduction to image entropy, correlation analysis of the RGB color space, and the log-Gabor filter. Section~\ref{sec:eva} provides an experimental analysis, and describes the testing and verification of the proposed method from multiple perspectives. Finally, Section~\ref{sec:conclusion} concludes with a summary of our work.

\section{The Proposed ENIQA Framework} \label{sec:proposed}
\begin{figure*}[htbp]
    \centering
    \includegraphics[width=0.98\textwidth]{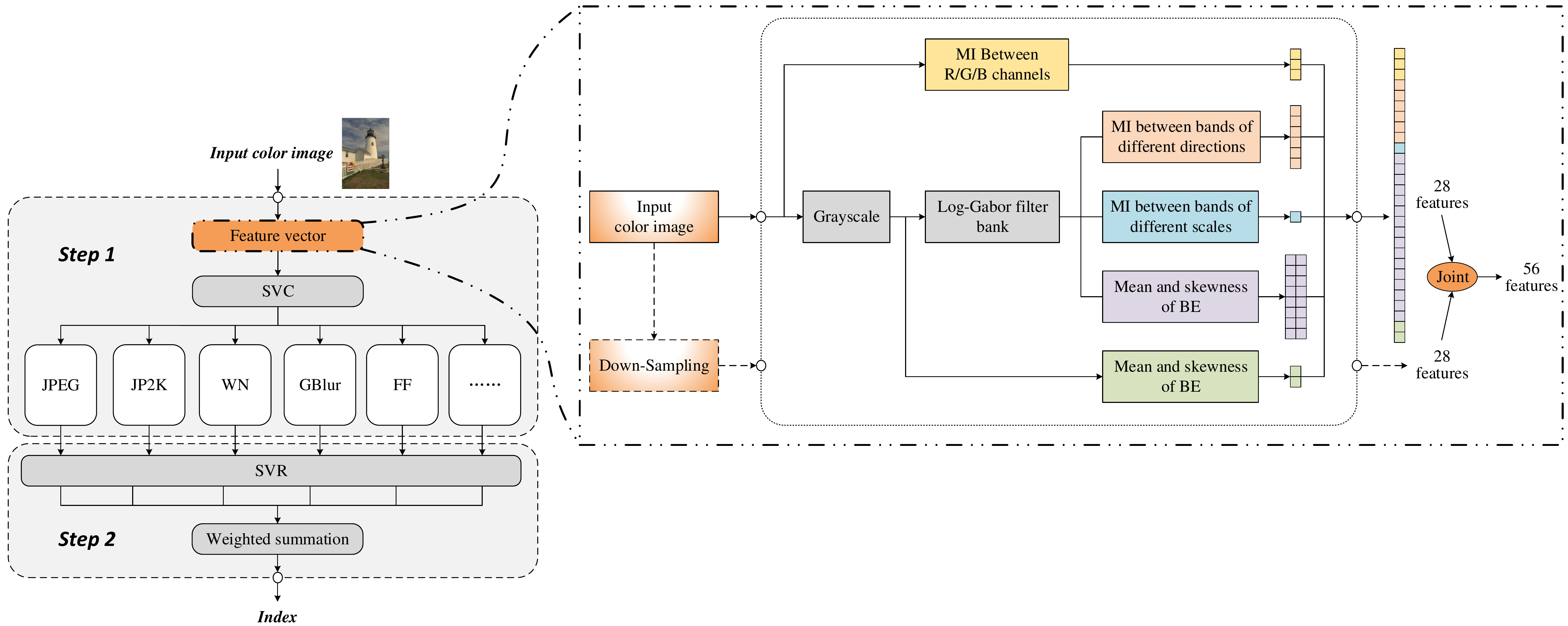}
    \caption{The framework of the proposed ENIQA method}
    \label{fig:illustrate}
\end{figure*}

In order to describe the local information of the image, the proposed ENIQA method introduces the MI and the TE in both the spatial and frequency domains. Given a color image whose quality is to be assessed, the MI between the three channels --— R, G, and B —-- is first calculated as feature group $1$. To extract feature group $2$, we convert the input image to grayscale and divide it into patches to calculate patch-wise entropy values. The obtained local entropy values are then pooled. The mean and the skewness then make up feature group $2$. For the frequency domain features, we apply log-Gabor filtering at two center frequencies and in four orientations to the grayscale image and obtain eight sub-band images, on which blocking and entropy calculation are implemented. The eight pairs of mean and skewness values are obtained from each sub-band, and they constitute feature group $3$. Furthermore, the MI between the sub-band images in the four different orientations and that between the two center frequencies are also calculated, respectively, as feature group $4$ and feature group $5$. The image is down-sampled using the nearest-neighbor method to capture multiscale behavior, yielding another set of $28$ features. Thus, ENIQA extracts a total of $56$ features for an input color image, as tabulated in Table~\ref{tab:features}. The right half of Fig.~\ref{fig:illustrate} illustrates the extraction process of the five feature groups.

After all the features are extracted, the proposed ENIQA method utilizes a two-stage framework to obtain a score index of the test image. In the first stage, the presence of a set of distortions in the image is estimated via SVC, giving the amount or probability of each type of distortion. In the second stage, for each type of distortion we consider, a support vector machine~\cite{burges1998tutorial} is trained to perform a regression that maps the features to the objective quality. Finally, the quality score of the image is produced by a weighted summation, where the probabilities from the first stage are multiplied by the corresponding regressed scores from the second stage and then added altogether. The left half of Fig.~\ref{fig:illustrate} shows the structure of the two-stage framework.

\begin{table*}[htbp]
  \centering
  \caption{Features used for ENIQA}
    \begin{tabular}{ccl}
    \hline
    \hline
    Group & Feature vector & \multicolumn{1}{c}{Feature description} \\
    \hline
    $1$     & $f_1-f_6$ & MI between the RGB channels for two scales \\
    $2$     & $f_7-f_{10}$ & Mean and skewness of the TE of the grayscale image for two scales \\
    $3$     & $f_{11}-f_{42}$ & Mean and skewness of the TE of the eight sub-band images for two scales \\
    $4$     & $f_{43}-f_{54}$ & MI of the sub-band images in different orientations for two scales  \\
    $5$     & $f_{55}-f_{56}$ & MI of the sub-band images at different center frequencies for two scales \\
    \hline
    \hline
    \end{tabular}%
  \label{tab:features}%
\end{table*}%

\subsection{Two-Dimensional Entropy}    \label{sec:TDEntropy}
Image entropy is a statistical feature that reflects the average information content in an image. The one-dimensional entropy of an image represents the information contained in the aggregated features of the grayscale distribution in the image, but does not contribute to the extraction of the spatial features. In order to characterize the local structure of the image, TE that describes the spatial correlation of the grayscale values is introduced.

After the color image $\bm{X}$ is converted to grayscale, the neighborhood mean of the grayscale image is selected as the spatial distribution feature. Let $p(x)$ denote the proportion of pixels whose gray value is $x$ in image $\bm{X}$, the one-dimensional entropy of a gray image is defined as:
\begin{equation}
  H_1(\bm{X}) = -\sum_{x=0}^{255}p(x)\log_2 p(x)
\end{equation}

The gray level of the current pixel and the neighborhood mean then form a feature pair, which is denoted as ($x_1$, $x_2$), where $x_1$ is the gray level of the pixel ($0 \leq  x_1 \leq  255$) and $x_2$ is the mean value of the neighbors ($0 \leq  x_2 \leq  255$). The combined probability density distribution function of $x_1$ and $x_2$ is given by:
\begin{equation}
    p(x_1, x_2)=\frac{f(x_1, x_2)}{MN}
\end{equation}
where $f(x_1, x_2)$ is the frequency at which the feature pair ($x_1$, $x_2$) appears, and the size of $\bm{X}$ is $M \times N$.

\begin{figure}[htbp]
    \centering
    \includegraphics[width=2.5in]{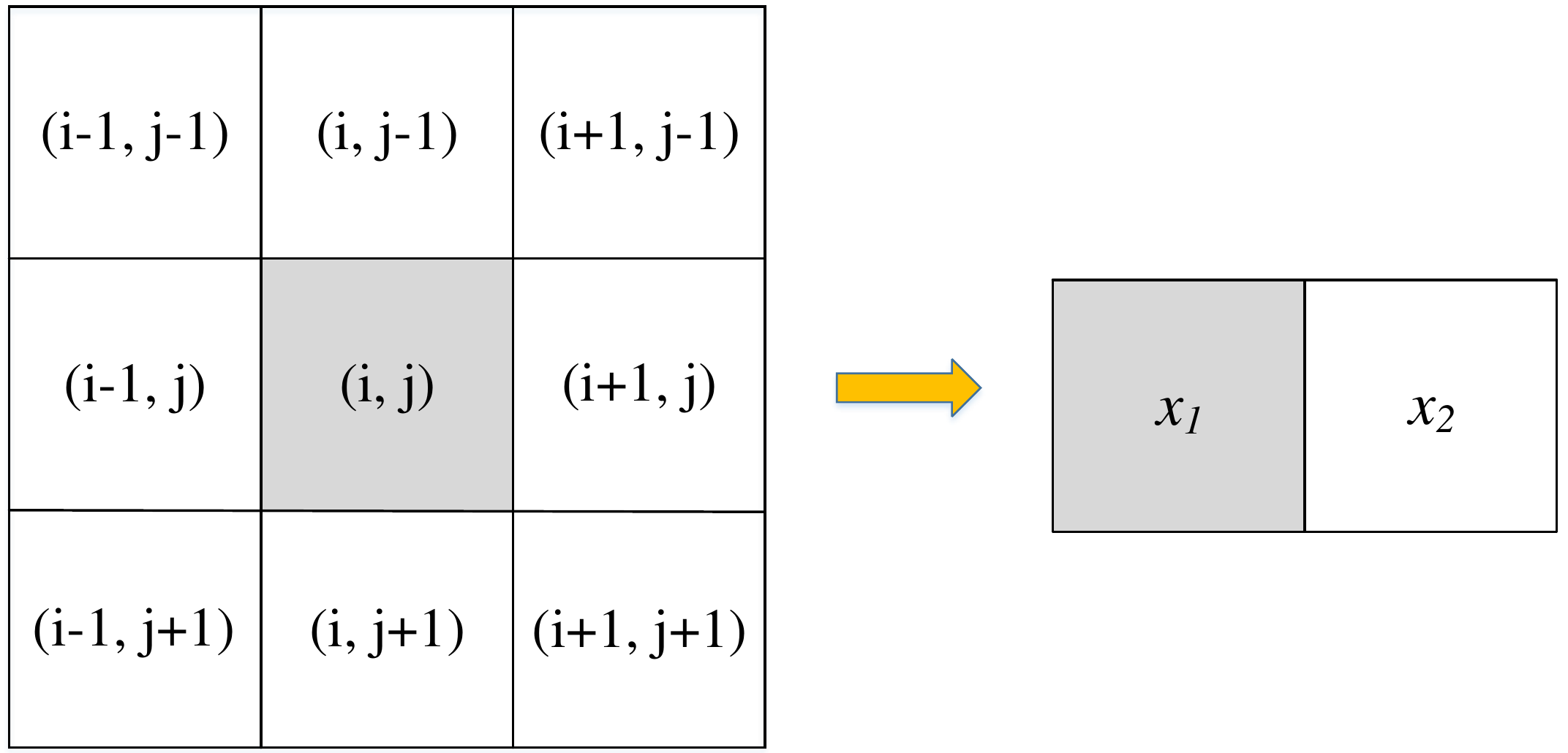}
    \caption{A pixel and its eight neighborhoods}
    \label{fig:area}
\end{figure}

In our implementation, $x_2$ is based on the eight adjacent neighbors of the center pixel, as shown in Fig.~\ref{fig:area}. The discrete TE is defined as:
\begin{equation}    \label{eq:disTE}
    H_2(\bm{X})=-\sum_{x_1=0}^{255}\sum_{x_2=0}^{255}p(x_1, x_2 ) \log_2 p(x_1, x_2)
\end{equation}

\begin{figure}[htbp]
    \centering
    \includegraphics[width=2.5in]{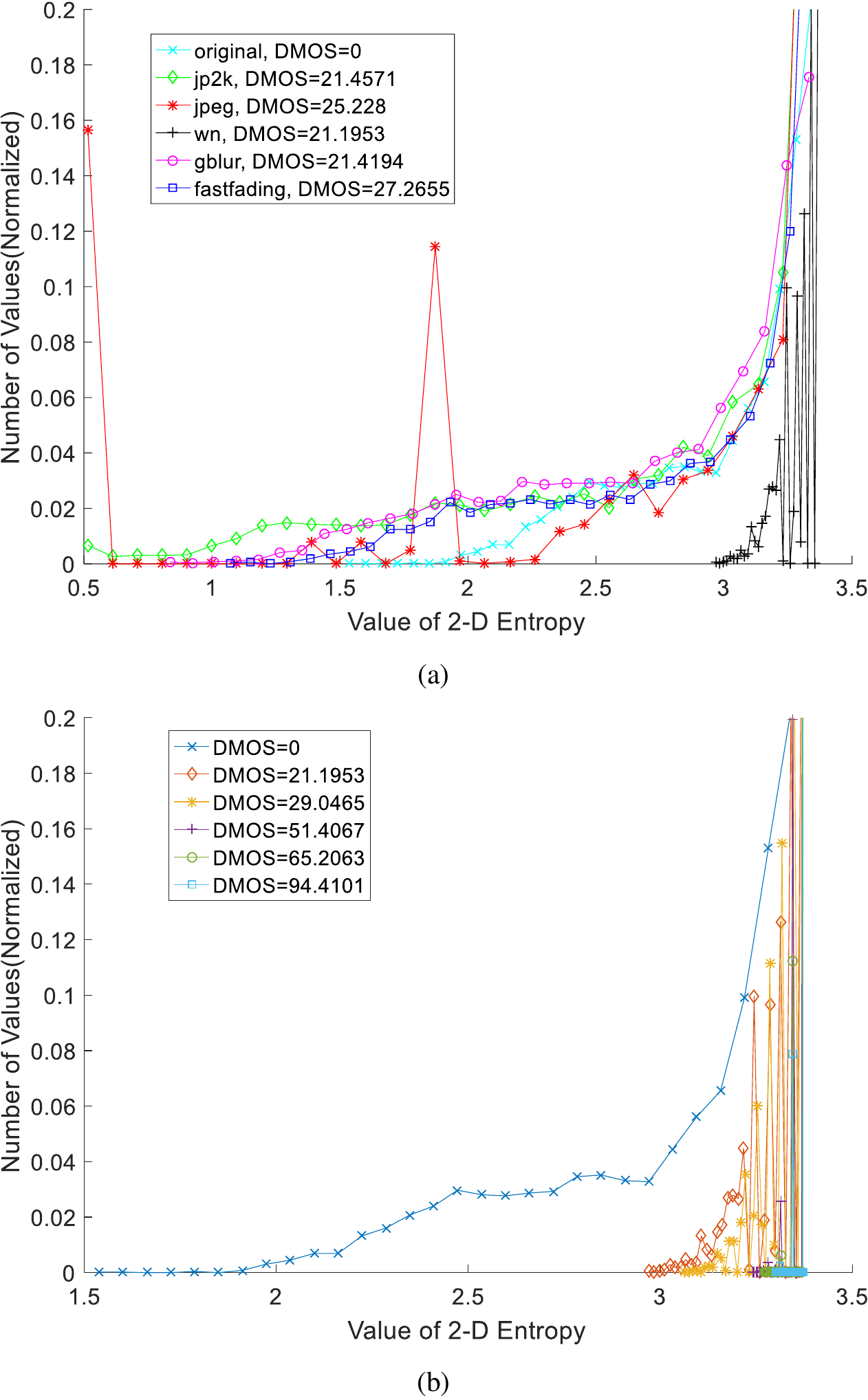}
    \caption{Histograms of TE values. (a) The six curves correspond to an undistorted image and its distorted counterparts with the same distortion level but different distortion types: “original” (DMOS=$0$), “JP2K” (DMOS=$21.4571$), “JPEG” (DMOS=$25.2228$), “WN” (DMOS=$21.1953$), “GBlur” (DMOS=$21.4194$), and “FF” (DMOS=$27.2655$). (b) The six curves correspond to an undistorted image and its distorted counterparts with the same distortion type but different distortion levels. The distortion type is GBlur and the DMOS values are $0$, $21.4194$, $41.6220$, $43.2646$, $55.7986$, and $83.2695$, respectively}
    \label{fig:statistics}
\end{figure}

The TE based on the above can describe the comprehensive features of the grayscale information of the pixel and the grayscale distribution in the neighborhood of the pixel. We determined the TE for a reference image ($monarch.bmp$ in the LIVE~\cite{sheikh2005live} database) and the five corresponding distorted images with the same distortion level but different distortion types. The statistical characteristics are shown in Fig.~\ref{fig:statistics}(a). All the differential mean opinion score (DMOS)~\cite{sheikh2006statistical} values are around $25$, and the distortion types span JPEG and JP2K compression, additive white Gaussian noise (WN), Gaussian blur (GBlur), and fast fading (FF) Rayleigh channel distortion. Similarly, the same experiment was also carried out on $monarch.bmp$ and the five corresponding distorted images with the same distortion type but different distortion levels (taking GBlur as an example), whose statistical characteristics are shown in Fig.~\ref{fig:statistics}(b). In Fig.~\ref{fig:statistics}, the abscissa axis represents the entropy and the vertical axis represents the normalized number of blocks. It can be seen from Fig.~\ref{fig:statistics} that both the distortion level and the distortion type can be distinguished by TE. Consequently, the TE can be considered a meaningful feature. Inspired by~\cite{ruderman1994statistics,liu2014no,liu2016blind}, we utilize the mean and skewness as the most typical features to describe the histogram.

The HVS automatically sets different priorities of attention for different regions of the observed image~\cite{zhang2016application}. Thus, before calculating the statistical characteristics of the TE, we conducted visual saliency detection on the image, $i.e.$, only the more important image patches were involved in the subsequent computation. To realize this, we first split the image into patches, pooled the patches according to human vision priority, and screened out the more significant ones. Then, according to the saliency values, we sorted the patches and calculated the mean and skewness of the local TE on the $80\%$ more important patches only. In the experiments, we used the spectral residual (SR) method~\cite{hou2007saliency} to generate the saliency map of the image to be measured. It is worth noting that the frequencies of different pixel values (integers from $0$ to $255$) are counted in every important patch to estimate the probability distributions in Eq.~(\ref{eq:disTE}).

\subsection{Mutual Information}    \label{sec:MInformation}
The application of colors in image display can not only stimulate the eye, but also allows the observer to perceive more information. The human eye has the ability to distinguish between thousands of colors, in spite of the perception of only dozens of gray levels~\cite{rose1957quantum}. There is a strong correlation between the RGB components of an image, which is embodied by the fact that the changes of individual color components reflected in the same region tend to be synchronized. That is to say, when the color of a certain area of a natural color image changes, the pixel gray values of the corresponding R, G, and B components also change at the same time. Moreover, although the gray value of a pixel varies with the color channels, different RGB components have quite good similarity and consistency in textures, edges, phases, and grayscale gradients~\cite{ren2010fusion}. Therefore, it is meaningful to characterize the MI between the three channels of R, G, and B.

Taking R and G as an example, it is assumed that $x_r$ and $x_g$ are the gray values of the red and green components of the input color image $\bm{X}$, while $p(x_r)$, $p(x_g)$ are the grayscale probability distribution functions in the two channels. $p(x_r, x_g)$ is the joint probability distribution function. The MI between the R and G channels is then formulated as:
\begin{equation}
  \begin{aligned}
    I(\bm{X}_R; \bm{X}_G ) &= H_1(\bm{X}_R ) + H_1(\bm{X}_G ) - H_2(\bm{X}_R, \bm{X}_G )     \\
                 &= \sum_{x_r=0}^{255}\sum_{x_g=0}^{255}p(x_r, x_g) \log_2 \frac{p(x_r, x_g)}{p(x_r) p(x_g)}
  \end{aligned}
\end{equation}
where $H_1 (\bm{X}_R )$ and $H_1 (\bm{X}_G )$ are the one-dimensional entropy of the corresponding channel, and $H_2 (\bm{X}_R, \bm{X}_G )$ represents the two-dimensional entropy between the two images, which is defined as:
\begin{equation}    \label{eq:JointEntropy}
    H_2(\bm{X}_R, \bm{X}_G)=-\sum_{x_r=0}^{255}\sum_{x_g=0}^{255}p(x_r, x_g ) \log_2 p(x_r, x_g)
\end{equation}

\subsection{Log-Gabor Filtering}    \label{sec:LGFiltering}
It is known that the log-Gabor filter function conforms to the HVS and is consistent with the symmetry of the cellular response of the human eye at logarithmic frequency scales~\cite{boukerroui2004choice}. The log-Gabor filter eliminates the DC component, overcomes the bandwidth limitation of the conventional Gabor filter, and has a typical frequency response with a Gaussian shape~\cite{field1987relations}. Thus, it is much easier, as well as more efficient, for a log-Gabor filter to extract information on a higher band. The transfer function of a two-dimensional log-Gabor filter can be expressed as:
\begin{equation}  \label{eq:logGaborfilter}
    G(f, \theta) = \exp\left ( -\frac{(\log(f/f_0))^2}{2(\log(\sigma_r/f_0))^2} \right ) \exp\left ( -\frac{(\theta - \theta_0)^2}{2\sigma_\theta^2} \right )
\end{equation}

In Eq.~\ref{eq:logGaborfilter}, $f_0$ gives the center frequency and $\theta_0$ represents the center orientation. $\sigma_r$ and $\sigma_\theta$ are the width parameters for the frequency and the orientation, respectively.

We distill the features in the frequency domain by implementing convolution on the log-Gabor filter and the image. The log-Gabor filter bank designed in this study consists of four filters, with orientations of $0^{\circ}$, $45^{\circ}$, $90^{\circ}$, and $135^{\circ}$, and two frequency bands. Eight sub-band images in four orientations and two bands are obtained after the input image is filtered.

\section{Experimental Results} \label{sec:eva}
In order to assess the statistical performance of the proposed method, we carried out experiments on the LIVE~\cite{sheikh2005live} and TID2013~\cite{ponomarenko2015image} databases. The LIVE database consists of $29$ reference images and $779$ distorted images of five distortion categories, while the TID2013 database contains $25$ reference images and $3000$ distorted images of $24$ distortion categories. Of these $25$ images, only $24$ are natural images, so we only used the $24$ natural images in the testing. At the same time, in order to ensure the consistency of the training and testing, we carried out the cross-database testing only over the four distortion categories in common with the LIVE database, namely, JP2K, JPEG, WN, and GBlur.

The indices used to measure the performance of the proposed method are the Spearman’s rank-order correlation coefficient (SROCC), the Pearson linear correlation coefficient (PLCC), and the root-mean-square error (RMSE) between the predicted DMOS and the ground-truth DMOS~\cite{corriveau2003final}. A value close to $1$ for SROCC and PLCC and a value close to $0$ for RMSE indicates better correlation with human perception. It is worth noting that PLCC and RMSE were computed after the predicted DMOS values were fitted by a nonlinear logistic regression function with five parameters~\cite{sheikh2006statistical}.
\begin{equation}
    f(z)=\beta _{1}\left [ \frac{1}{2}-\frac{1}{1+exp(\beta _{2}(z-\beta _{3}))} \right ]+\beta _{4}z+\beta _{5}
\end{equation}
where $z$ is the objective IQA score, $f(z)$ is the IQA regression fitting score, and $\beta _{i}(i=1,2,\cdots ,5)$ are the parameters of the regression function.

\subsection{Correlation of Feature Vectors with Human Opinion}    \label{sec:CFVHOpinion}
\begin{figure}[htbp]
    \centering
    \includegraphics[width=3.0in]{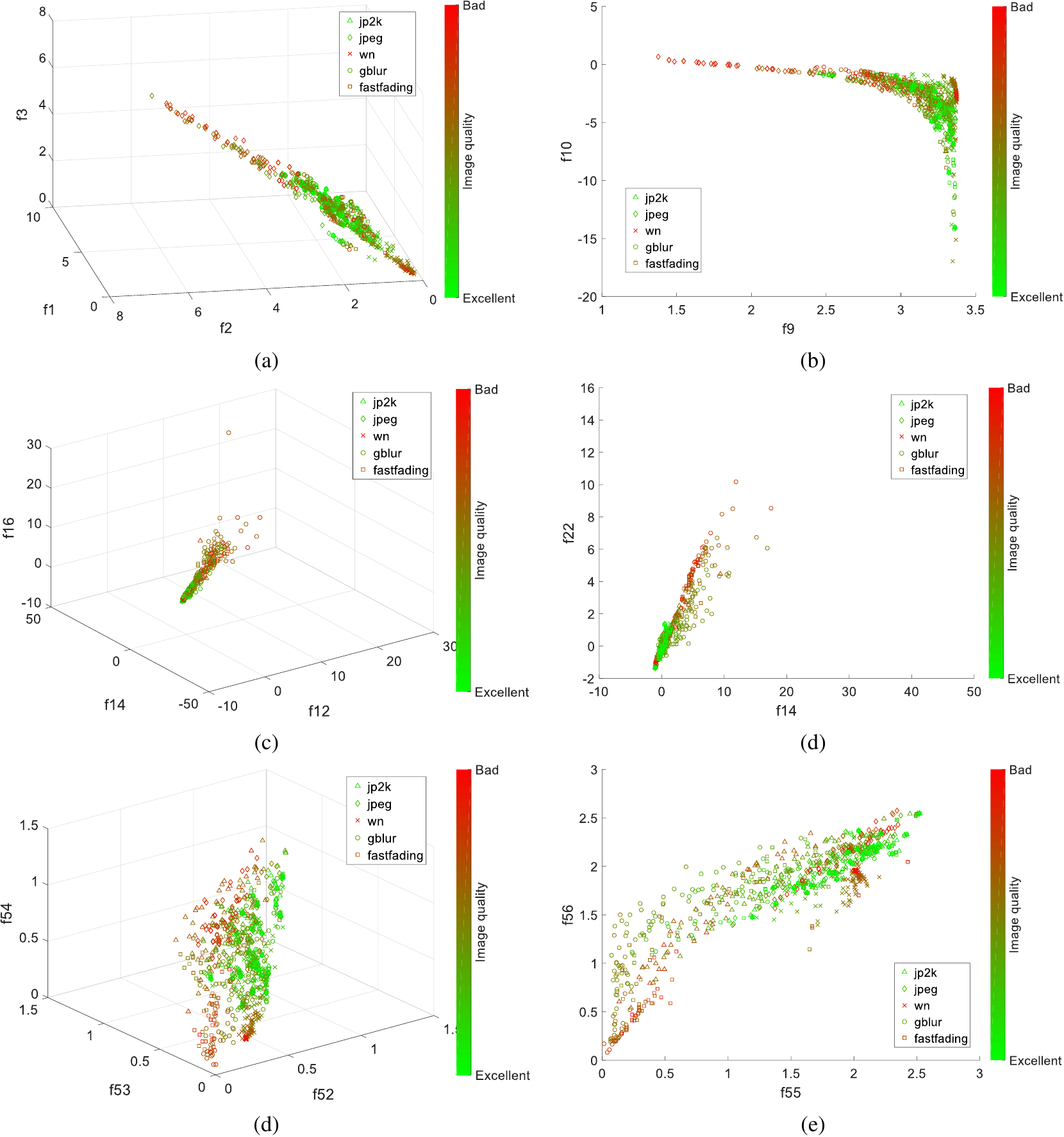}
    \caption{Illustration of the discriminatory power of different feature combinations (zoom in to get the markers more discriminative). (a) Elements $1$, $2$, and $3$. (b) Elements $9$ and $10$. (c) Elements $12$, $14$, and $16$ (d). Elements $14$ and $22$. (e) Elements $52$, $53$, and $54$. (f) Elements $55$ and $56$}
    \label{fig:distinguish}
\end{figure}

In this experiment, we assessed the discriminatory power of different feature combinations. With the feature groups listed in Table~\ref{tab:features}, we visually illustrate the relationship between image quality and features in the form of two-dimensional$/$three-dimensional scatter plots. As shown in Fig.~\ref{fig:distinguish}, the different feature combinations are used as the axes, and each image in the LIVE database corresponds to a scatter point in the coordinate system. Furthermore, we use different markings to distinguish the five types of distortion and map the score of each image to the preset colormap. The ideal case is that the points with different distortion types are well separated. In this paper, we selected only a few representative images as examples. It can be seen from Fig.~\ref{fig:distinguish}(a) and \ref{fig:distinguish}(b) that the scatter points of JPEG and WN have a very different spatial distribution than the other points, which allows them to be better distinguished. From Fig.~\ref{fig:distinguish}(c) and \ref{fig:distinguish}(d), we can see that GBlur can be distinguished, to some extent, from the other types of distortion. However, for GBlur points with lower distortion levels, they cannot be easily separated from FF and JP2K, since the distributions of the scatter points of these three distortion types are very similar. As can be observed in Fig.~\ref{fig:distinguish}(e) and \ref{fig:distinguish}(f), images with higher distortion levels of WN, GBlur, and FF are more easily distinguished from images with good quality. Nonetheless, GBlur and FF are indistinguishable. And still, JP2K points cause the reduction of distinguishability, as some of them are scattered close to the highly-distorted GBlur and FF points. According to Fig.~\ref{fig:distinguish}, the number of features we selected seems too small to distinguish all the distortion types. Due to the limitation of human spatial cognition, it is difficult for us to show the discriminative ability of the features in a graphical way, such as a four-dimensional scatter plot, by selecting feature combinations of a higher dimension. In Section~\ref{sec:CPAnalysis}, we prove that when more features are selected (actually, we chose 56-dimensional features), the discriminatory power of the feature vector on the distortion type is further enhanced, which indicates the accuracy and reliability of our selection of features.

\subsection{Correlation of Individual Feature Vectors with Human Perception}    \label{sec:CIFVHPerception}
In order to quantitatively study the predictive ability of each feature vector, we performed a recombination of the features in Table~\ref{tab:features}, separately deployed specific subsets (feature vectors), and designed three limited models: $1)$ The feature vector $f_1–f_6$ represents the MI between the three color channels on two scales, denoted as $ENIQA_1$. $2)$ The feature vector $f_7–f_{42}$ represents the mean and skewness of the TE on two scales, denoted as $ENIQA_2$. $3)$ The feature vector $f_{43}–f_{56}$ represents the MI between the sub-band images on two scales, denoted as $ENIQA_3$.

We performed the assessment of these three limited models by $1000$ train-test iterations of cross-validation. In each iteration, we randomly split the LIVE~\cite{sheikh2005live} database into two non-overlapping sets: a training set comprising $80\%$ of the reference images as well as their corresponding distorted counterparts, and a test set composed of the remaining $20\%$. Finally, the median SROCC, PLCC, and RMSE values over $1000$ trials are reported as the final performance indices, as shown in Table~\ref{tab:msroccind}$-$\ref{tab:mrmseind}. It is not difficult to see that each feature vector has a different degree of correlation with the subjective assessment. Among them, the TE contributes the most to the performance of the method, followed by the MI between the sub-band images. Although the MI between the color channels contributes the least, it is a valuable extension of the TE feature.
\begin{table}[htbp]
  \centering
  \caption{Median SROCC values across $1000$ train-test trials on the LIVE database}
  \scalebox{0.8}{
    \begin{tabular}{ccccccc}
    \hline
    \hline
    Model & JP2K  & JPEG  & WN & GBlur  & FF    & All \\
    \hline
    $ENIQA_{1}$ & 0.3209  &  0.8366  &  0.9526  &  0.1994  &  0.3218  &  0.5704 \\
    $ENIQA_{2}$ & 0.8784  &  0.9501  &  0.9589  &  0.9198  &  0.8168  &  0.9054 \\
    $ENIQA_{3}$ & 0.9058  &  0.8235  &  0.9580  &  0.9293  &  0.8390  &  0.8540 \\
    \hline
    \hline
    \end{tabular}}%
  \label{tab:msroccind}%
\end{table}%

\begin{table}[htbp]
  \centering
  \caption{Median PLCC values across $1000$ train-test trials on the LIVE database}
  \scalebox{0.8}{
    \begin{tabular}{ccccccc}
    \hline
    \hline
    Model & JP2K  & JPEG  & WN & GBlur  & FF    & All \\
    \hline
    $ENIQA_{1}$ & 0.5052  &  0.8974  &  0.9565  &  0.5557  &  0.4989  &  0.6663 \\
    $ENIQA_{2}$ & 0.9197  &  0.9769  &  0.9726  &  0.9248  &  0.8701  &  0.9218 \\
    $ENIQA_{3}$ & 0.9397  &  0.8362  &  0.9697  &  0.9242  &  0.8672  &  0.8474 \\
    \hline
    \hline
    \end{tabular}}%
  \label{tab:mplccind}%
\end{table}%

\begin{table}[htbp]
  \centering
  \caption{Median RMSE values across $1000$ train-test trials on the LIVE database}
  \scalebox{0.8}{
    \begin{tabular}{ccccccc}
    \hline
    \hline
    Model & JP2K  & JPEG  & WN & GBlur  & FF & All \\
    \hline
    $ENIQA_{1}$ & 24.8268  &  15.4386  &  9.5981  &  19.9254  &  26.6129  &  23.2067 \\
    $ENIQA_{2}$ & 11.4210  &  7.4958  &  7.6829  &  9.0478  &  14.9870  &  12.0568 \\
    $ENIQA_{3}$ & 9.8370  &  19.2657  &  8.0514  &  9.1473  &  15.3000  &  16.5505 \\
    \hline
    \hline
    \end{tabular}}%
  \label{tab:mrmseind}%
\end{table}%

\subsection{Comparison with Other IQA Methods}    \label{sec:COMethods}
To further illustrate the superiority of the proposed method, we compared ENIQA with $10$ other state-of-the-art IQA methods. The three FR-IQA approaches were the peak signal-to-noise ratio (PSNR), the structural similarity index (SSIM)~\cite{wang2004image}, and visual information fidelity (VIF)~\cite{sheikh2004image}, and the seven NR-IQA approaches were BIQI~\cite{moorthy2010two}, DIIVINE~\cite{moorthy2011blind}, BLIINDS-II~\cite{saad2012blind}, BRISQUE~\cite{mittal2012no}, NIQE~\cite{mittal2013making}, ILNIQE~\cite{zhang2015feature}, and SSEQ~\cite{liu2014no}. To make a fair comparison, we used the same $80\%$ training$/$$20\%$ testing protocol over $1000$ iterations on all the models. The source code of all the methods was provided by the authors. In the training of an NR model, the LIBSVM toolkit~\cite{chang2011libsvm} was used to implement SVC and SVR, both adopting a radial-basis function (RBF) kernel. We selected an $e-SVR$ model for regression, and both the cost and the $\gamma $ for RBF are set to $1e^{-4}$. Since the FR approaches do not require a training procedure, they were only performed on distorted images, $i.e.$, the reference images were not included. For the results listed in Table~\ref{tab:msrocciqa}$-$\ref{tab:mrmseiqa}, the top performances in the FR-IQA indices and those in the NR-IQA indices are highlighted in bold. For the NR-IQA indices, we have also underlined the second-best results.

It can be seen that the proposed ENIQA method performs well on the LIVE database. To be specific, ENIQA obtains the highest SROCC value for JPEG and WN, and the second-highest overall SROCC value among the NR methods listed in Table~\ref{tab:msrocciqa}. In terms of PLCC and RMSE, ENIQA is superior to all the other NR methods, except BRISQUE, on JPEG and WN, and also ranks second in overall performance. Generally speaking, the overall performance of the proposed ENIQA method is superior to most of the other NR methods, and is even ahead of some of the classic FR methods such as SSIM. Besides, ENIQA is rather good at evaluating images with distortions of JPEG and WN.
\begin{table}[htbp]
  \centering
  \caption{Median SROCC values on the LIVE database}
  \scalebox{0.75}{
    \begin{tabular}{ccccccc}
    \hline
    \hline
    Method & JP2K  & JPEG  & WN & GBlur  & FF    & All \\
    \hline
    PSNR  & 0.9053 & 0.8866 & \textbf{0.9844} & 0.8120 & 0.8981 & 0.8850 \\
    SSIM  & \textbf{0.9517} & \textbf{0.9671} & 0.9631 & 0.9306 &0.9404 & 0.9255 \\
    VIF   & 0.9160 & 0.9482 & 0.9435 & \textbf{0.9600}  & \textbf{0.9617} & \textbf{0.9287} \\
    \hline
    \textit{BIQI} & 0.8401 & 0.8425 & 0.9362 & 0.8924 & 0.7383 & 0.8340 \\
    \textit{DIIVINE} & \underline{0.9363} & 0.9051 & 0.9692 & 0.9478 & 0.8778 & 0.9261 \\
    \textit{BLIINDS-II} & \textbf{0.9389} & 0.9449 & 0.9596 & 0.9447 & 0.8653 & 0.9362 \\
    \textit{BRISQUE} & 0.9349 & 0.9480 & 0.9725 & \underline{0.9616} & 0.8821 & \textbf{0.9411} \\
    \textit{NIQE} & 0.9171 & 0.9094 & 0.9697 & \textbf{0.9678} & 0.8715 & 0.9142 \\
    \textit{ILNIQE} & 0.9199 & 0.9335 & \underline{0.9730} & 0.9526 & \textbf{0.8991} & 0.9219 \\
    \textit{SSEQ} & 0.9355 & \underline{0.9509} & 0.9689 & 0.9554 & \underline{0.8943} & 0.9349 \\
    \textit{ENIQA} & 0.9255 & \textbf{0.9515} & \textbf{0.9762} & 0.9481 & 0.8491 & \underline{0.9378} \\
    \hline
    \hline
    \end{tabular}}%
  \label{tab:msrocciqa}%
\end{table}%

\begin{table}[htbp]
  \centering
  \caption{Median PLCC values on the LIVE database}
  \scalebox{0.75}{
    \begin{tabular}{ccccccc}
    \hline
    \hline
    Method & JP2K  & JPEG  & WN & GBlur  & FF    & All \\
    \hline
    PSNR  & 0.9176 & 0.9130 & \textbf{0.9887} & 0.8277 & 0.9169 & 0.8825 \\
    SSIM  & \textbf{0.9508} & \textbf{0.9751} & 0.9783 & 0.8739 & 0.9203 & 0.9241 \\
    VIF   & 0.9360 & 0.9594 & 0.9679 & \textbf{0.9689} & \textbf{0.9748} & \textbf{0.9318} \\
    \hline
    \textit{BIQI} & 0.8778 & 0.8827 & 0.9551 & 0.8873 & 0.7987 & 0.8494 \\
    \textit{DIIVINE} & 0.9541 & 0.9416 & 0.9791 & 0.9443 & 0.8994 & 0.9309 \\
    \textit{BLIINDS-II} & \textbf{0.9564} & 0.9721 & 0.9698 & 0.9533 & 0.8855 & 0.9412 \\
    \textit{BRISQUE} & \underline{0.9550} & \textbf{0.9789} & \textbf{0.9838} & 0.9601 & 0.9151 & \textbf{0.9510} \\
    \textit{NIQE} & 0.9485 & 0.9443 & 0.9396 & \textbf{0.9699} & 0.9040 & 0.8734 \\
    \textit{ILNIQE} & 0.9549 & 0.9670 & 0.9766 & 0.9487 & \textbf{0.9250} & 0.8987 \\
    \textit{SSEQ} & 0.9514 & 0.9638 & 0.9823 & \underline{0.9691} & \underline{0.9227} & 0.9380 \\
    \textit{ENIQA} & 0.9503 & \underline{0.9741} & \underline{0.9828} & 0.9447 & 0.8791 & \underline{0.9437} \\
    \hline
    \hline
    \end{tabular}}%
  \label{tab:mplcciqa}%
\end{table}%

\begin{table}[htbp]
  \centering
  \caption{Median RMSE values on the LIVE database}
  \scalebox{0.75}{
    \begin{tabular}{ccccccc}
    \hline
    \hline
    Method & JP2K  & JPEG  & WN & GBlur  & FF    & All \\
    \hline
    PSNR  & 9.9129 & 12.8262 & \textbf{4.2127} & 10.3076 & 11.1624 & 12.7229 \\
    SSIM  & \textbf{7.7506} & \textbf{6.9422} & 5.7534 & 9.1163 & 10.8487 & 10.3565 \\
    VIF   & 8.8859 & 8.8635 & 7.0133 & \textbf{4.527} & \textbf{6.2119} & \textbf{9.8436} \\
    \hline
    \textit{BIQI} & 13.7600 & 16.4621 & 9.7859 & 11.0204 & 18.7142 & 16.4823 \\
    \textit{DIIVINE} & 8.5666 & 11.8295 & 6.7472 & 7.9227 & 13.6016 & 11.4397 \\
    \textit{BLIINDS-II} & \textbf{8.1849} & 8.2528 & 8.0538 & 7.1483 & 14.051 & 10.5832 \\
    \textit{BRISQUE} & \underline{8.5280} & \textbf{7.1899} & \textbf{5.9282} & 6.6549 & 12.3081 & \textbf{9.6662} \\
    \textit{NIQE} & 9.1488 & 11.5084 & 11.4903 & \textbf{5.787} & 13.2733 & 15.1785 \\
    \textit{ILNIQE} & 8.5310 & 8.9270 & 7.0934 & 7.5084 & \textbf{11.7819} & 13.6439 \\
    \textit{SSEQ} & 8.8145 & 9.3283 & 6.8836 & \underline{6.5752} & \underline{12.2893} & 10.6405 \\
    \textit{ENIQA} & 8.9964 & \underline{7.9640} & \underline{6.1051} & 7.7741 & 14.5150 & \underline{10.3234} \\
    \hline
    \hline
    \end{tabular}}%
  \label{tab:mrmseiqa}%
\end{table}%

\subsection{Variation with Window Size}    \label{sec:VWSize}
As mentioned above, since the local saliency difference of the image is considered, the proposed ENIQA method blocks the image with a window and counts the frequency of the gray values in each block to generate feature pairs before calculating the local TE. Table~\ref{tab:winsizes} shows the effect of different window sizes on the performance of the proposed method, where the highest SROCC value of each column is highlighted in bold. The average time consumption for evaluating a single image is also reported in Table~\ref{tab:winsizes}. All the experiments are performed on a PC with Intel-i7-6700K CPU@4.0GHz, 16G RAM, MATLAB $R2016a$. The elapsed time is the mean value measured through $10$ times of evaluations on the same $384 \times 512 \times 3$ image.
\begin{table}[htbp]
  \centering
  \caption{Median SROCC value of ENIQA on the LIVE database with different window sizes}
  \scalebox{0.75}{
    \begin{tabular}{c|cccccc|c}
    \hline
    \hline
    \multirow{2}{*}{K, L} & \multicolumn{6}{c|}{SROCC}  & \multirow{2}{*}{Time(s)} \\
    \cline{2-7}          & JP2K  & JPEG  & WN & GBlur  & FF    & ALL   &  \\
    \hline
    6, 6   & 0.9102  &  0.9122  &  0.9728  &  0.9408  &  0.8468  &  0.9199  &  9.8541 \\
    8, 8   & \textbf{0.9255}  &  \textbf{0.9515}  &  \textbf{0.9762}  &  \textbf{0.9481}  &  \textbf{0.8491}  &  \textbf{0.9378}  &  4.4856 \\
    12, 12 & 0.9065  &  0.9058  &  0.9656  &  0.9402  &  0.8257  &  0.9143  &  3.0859 \\
    16, 16 & 0.9157  &  0.9137  &  0.9673  &  0.9368  &  0.8445  &  0.9199  &  2.0547 \\
    32, 32 & 0.9017  &  0.9021  &  0.9672  &  0.9207  &  0.8341  &  0.9120  &  2.4387 \\
    \hline
    \hline
    \end{tabular}}%
  \label{tab:winsizes}%
\end{table}%

\begin{figure}[htbp]
    \centering
    \includegraphics[width=2.5in]{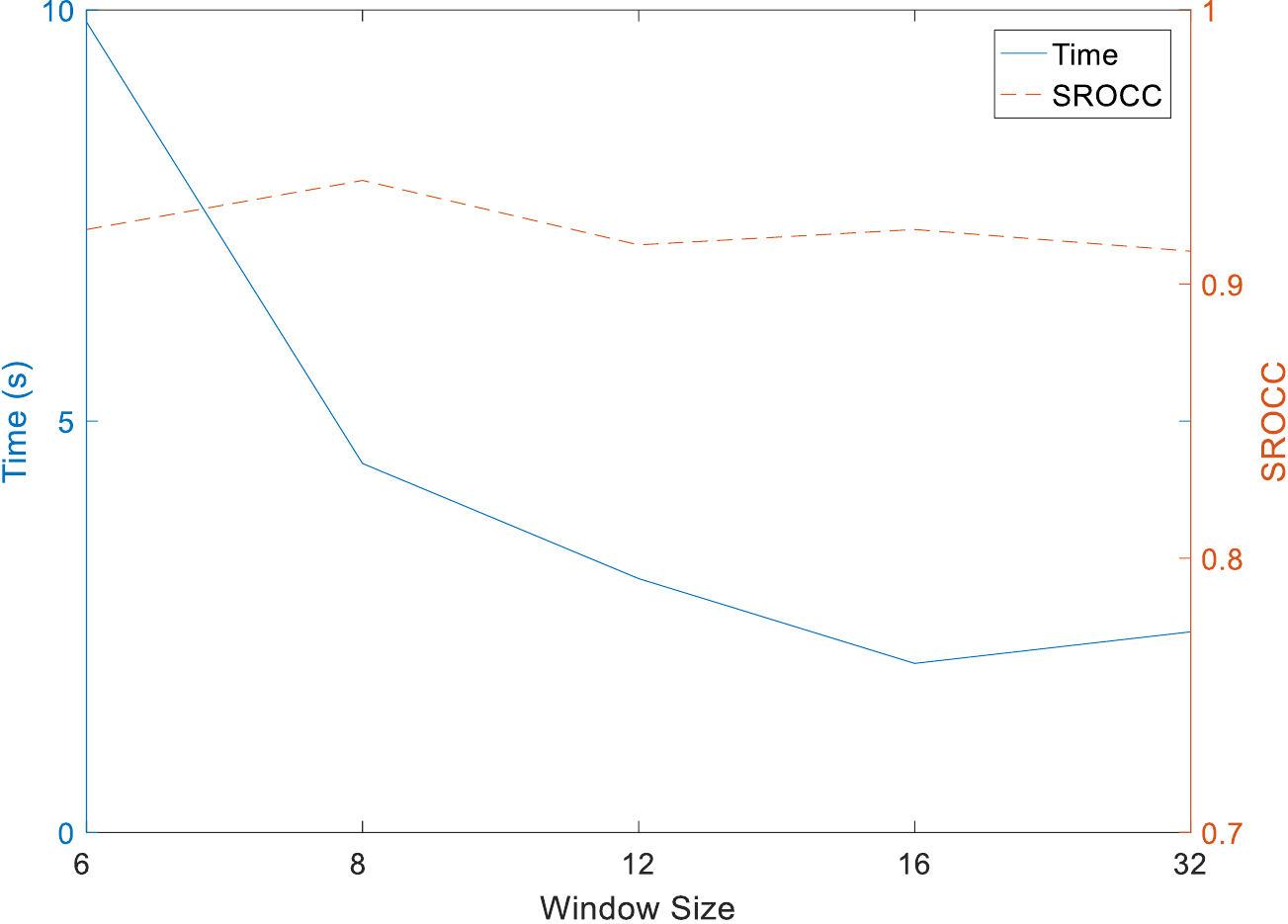}
    \caption{Line charts between the selected window size and the SROCC value as well as the average time consumed on evaluating a single image according to Table~\ref{tab:winsizes}. When the window size is set to $8 \times 8$, the method achieves best SROCC performance}
    \label{fig:SroccTime}
\end{figure}

In order to visualize the trend, we also drew two line charts in Fig.~\ref{fig:SroccTime}, which intuitively illustrate the change of the elapsed time and the SROCC value with the selected window size.

It can be observed that the performance of the proposed method varies with the size of the window. As the window size increases, the SROCC value shows a trend of increasing first and then decreasing, reaching a peak at $8 \times 8$. At the same time, the runtime of the method mostly decreases monotonically with the increase of the window size. To make a compromise, we used $K = L = 8$ in this study. It should be pointed out that the overall SROCC value is still maintained above $0.9$ when the window size is $32 \times 32$, which implies that the window size can be appropriately increased to trade accuracy for real-time performance in time-critical applications.

\subsection{Statistical Significance Testing}    \label{sec:SSTesting}
\begin{figure}[htbp]
    \centering
    \includegraphics[width=2.8in]{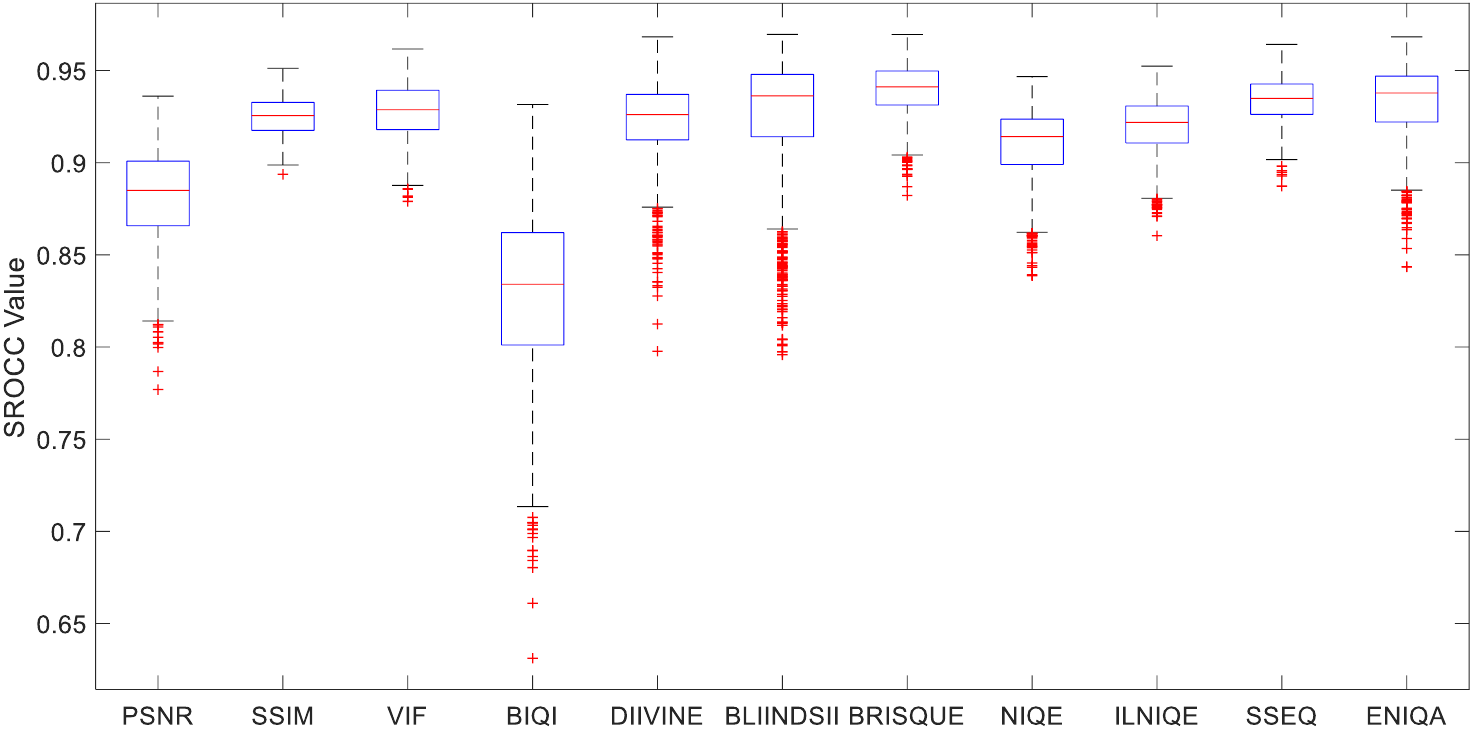}
    \caption{Box plot of SROCC distributions of the compared IQA methods over 1000 trials on the LIVE database}
    \label{fig:boxplot}
\end{figure}

In order to compare the performance of the different methods in a more intuitive way, Fig.~\ref{fig:boxplot} shows a box plot of the SROCC distributions for the $11$ IQA methods (including the proposed ENIQA method) across $1000$ train-test trials, which provides key information about the location and dispersion of the data. Meanwhile, we performed a two-sample t-test~\cite{Sheskin2003handbook} between the methods, and the results are shown in Table~\ref{tab:ttests}. The null hypothesis is that the mean correlation value of the row is equal to the mean correlation value of the column at the $95\%$ confidence level. The alternative hypothesis is that the mean correlation value of the row is greater (or less) than the mean correlation value of the column. Table~\ref{tab:ttests} indicates which row is statistically superior ('$1$'), statistically equivalent ('$0$'), or statistically inferior ('$-1$') to which column. Although BRISQUE and SSEQ are statistically superior to ENIQA in Table~\ref{tab:ttests}, it can be seen from Fig.~\ref{fig:boxplot} that ENIQA outperforms all the other FR and NR approaches, except BRISQUE, in terms of the median value.

\begin{table*}[htbp]
  \centering
  \caption{Results of the t-tests performed between SROCC values}
    \scalebox{0.85}{
    \begin{tabular}{cccccccccccc}
    \hline
    \hline
    Method & PSNR  & SSIM  & VIF   & BIQI  & DIIVINE & BLIINDS-II & BRISQUE & NIQE  & ILNIQE & SSEQ  & ENIQA \\
    \hline
    PSNR  & 0     & -1    & -1    & 1     & -1    & -1    & -1    & -1    & -1    & -1    & -1 \\
    SSIM  & 1     & 0     & -1    & 1     & 1     & 1     & -1    & 1     & 1     & -1    & -1 \\
    VIF   & 1     & 1     & 0     & 1     & 1     & 1     & -1    & 1     & 1     & -1    & -1 \\
    BIQI  & -1    & -1    & -1    & 0     & -1    & -1    & -1    & -1    & -1    & -1    & -1 \\
    DIIVINE & 1     & -1    & -1    & 1     & 0     & 1     & -1    & 1     & 1     & -1    & -1 \\
    BLIINDS-II & 1     & -1    & -1    & 1     & -1    & 0     & -1    & 1     & 1     & -1    & -1 \\
    BRISQUE & 1     & 1     & 1     & 1     & 1     & 1     & 0     & 1     & 1     & 1     & 1 \\
    NIQE  & 1     & -1    & -1    & 1     & -1    & -1    & -1    & 0     & -1    & -1    & -1 \\
    ILNIQE & 1     & -1    & -1    & 1     & -1    & -1    & -1    & 1     & 0     & -1    & -1 \\
    SSEQ  & 1     & 1     & 1     & 1     & 1     & 1     & -1    & 1     & 1     & 0     & 1 \\
    ENIQA & 1     & 1     & 1     & 1     & 1     & 1     & -1    & 1     & 1     & -1     & 0 \\
    \hline
    \hline
    \end{tabular}}%
  \label{tab:ttests}%
\end{table*}%

\subsection{Classification Performance Analysis}    \label{sec:CPAnalysis}
\begin{figure}[htbp]
    \centering
    \includegraphics[width=2.8in]{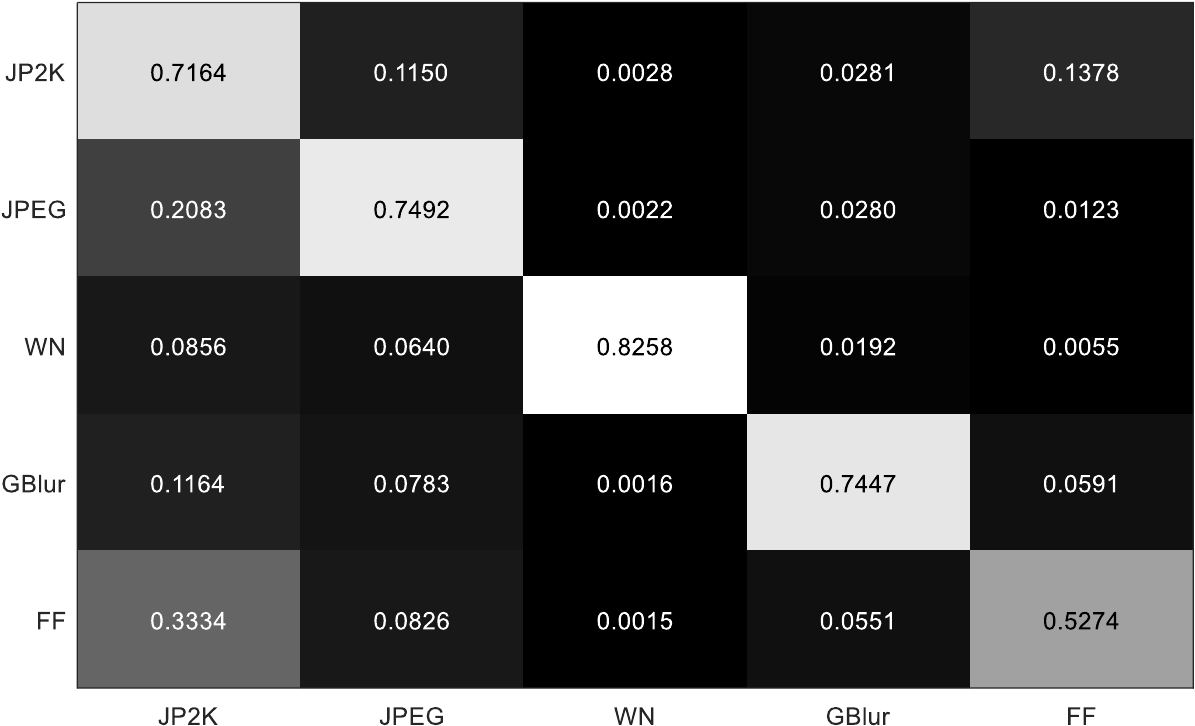}
    \caption{Mean confusion matrix of the classification accuracy across 1000 train-test trials}
    \label{fig:matrix}
\end{figure}

We analyzed the classification accuracy of ENIQA on the LIVE database based on the two-stage framework. The average classification accuracies for all the distortion types across $1000$ random trials are listed in Table~\ref{tab:classaccu}. It can be seen from Table~\ref{tab:classaccu} that when the feature dimensions reach $56$, the classification accuracy of JP2K reaches $71.6369\%$, which is fairly acceptable. In Section~\ref{sec:CFVHOpinion}, however, we showed that it is extremely difficult to distinguish JP2K images by low-dimensional feature vectors. Thus, we can speculate that in the 56-dimensional space composed of the features, the distorted images of the JP2K type are discernible by the hyperplane constructed by SVC. Furthermore, in order to visualize which distortion categories may be confused with each other, we plotted a confusion matrix~\cite{congalton1991review}, as shown in Fig.~\ref{fig:matrix}. Each value in the confusion matrix indicates the probability of the distortion category on the vertical axis being confused with that on the horizontal axis. The numerical values are the average classification accuracies of the $1000$ random trials.

It can be seen from Table~\ref{tab:classaccu} and Fig.~\ref{fig:matrix} that WN cannot easily be confused with the other distortion categories, while the other four distortion categories are more easily confused. As FF consists of JP2K followed by packet loss, it is understandable that FF distortion is more easily confused with JP2K compression distortion. From Fig.~\ref{fig:statistics}, we can also see that the TE distributions of WN and JPEG are very specific, while JP2K, GBlur, and FF have quite similar TE distributions, which results in them being more easily confused.
\begin{table}[htbp]
  \centering
  \caption{Mean classification accuracy across 1000 train-test trials}
  \scalebox{0.75}{
    \begin{tabular}{ccccccc}
    \hline
    \hline
          & JP2K  & JPEG  & WN & GBlur  & FF    & All \\
    \hline
    Class.Acc(\%) & 71.6369  &  74.9191  &  82.5750  &  74.4694  &  52.7417  &  71.2684 \\
    \hline
    \hline
    \end{tabular}}%
  \label{tab:classaccu}%
\end{table}%

\subsection{Database Independence}    \label{sec:DIndependence}
In order to test the generalization ability of the assessment model to different samples, we trained the model on the whole LIVE database and tested it on the TID2013 database, noting that we only chose distortion types in common with the LIVE database (JP2K, JPEG, WN, and GBlur). The computed performance indices are shown in Table~\ref{tab:crossdb}, and the top performances for the FR-IQA indices and those for the NR-IQA indices are highlighted in bold. For the NR-IQA indices, we have also underlined the second-best results. It is clear that the proposed ENIQA method remains competitive on TID2013, with a superior performance to all the other NR methods, including BRISQUE, which shows an excellent performance on the LIVE database.
\begin{table}[htbp]
  \centering
  \caption{Performance indices obtained by training on the LIVE database and testing on the TID2013 database}
  \scalebox{0.85}{
    \begin{tabular}{cccc}
    \hline
    \hline
    Method & SROCC & PLCC  & RMSE \\
    PSNR  & \textbf{0.9244} & 0.9140 & 0.5671 \\
    SSIM  & 0.8662 & 0.8723 & 0.6836 \\
    VIF   & 0.9181 & \textbf{0.9401} & \textbf{0.4765} \\
    \hline
    \textit{BIQI} & 0.8388 & 0.8447 & 0.7482 \\
    \textit{DIIVINE} & 0.7958 & 0.7868 & 0.8628 \\
    \textit{BLIINDS-II} & 0.8503 & 0.8413 & 0.7556 \\
    \textit{BRISQUE} & \underline{0.8817} & \underline{0.8860} & \underline{0.6482} \\
    \textit{NIQE} & 0.6798 & 0.6685 & 1.0397 \\
    \textit{ILNIQE} & 0.7793 & 0.7956 & 0.8469 \\
    \textit{SSEQ} & 0.8294 & 0.8769 & 0.6719 \\
    \textit{ENIQA} & \textbf{0.8973} & \textbf{0.9009} & \textbf{0.6067} \\
    \hline
    \hline
    \end{tabular}}%
  \label{tab:crossdb}%
\end{table}%

\begin{figure}[htbp]
    \centering
    \includegraphics[width=2.2in]{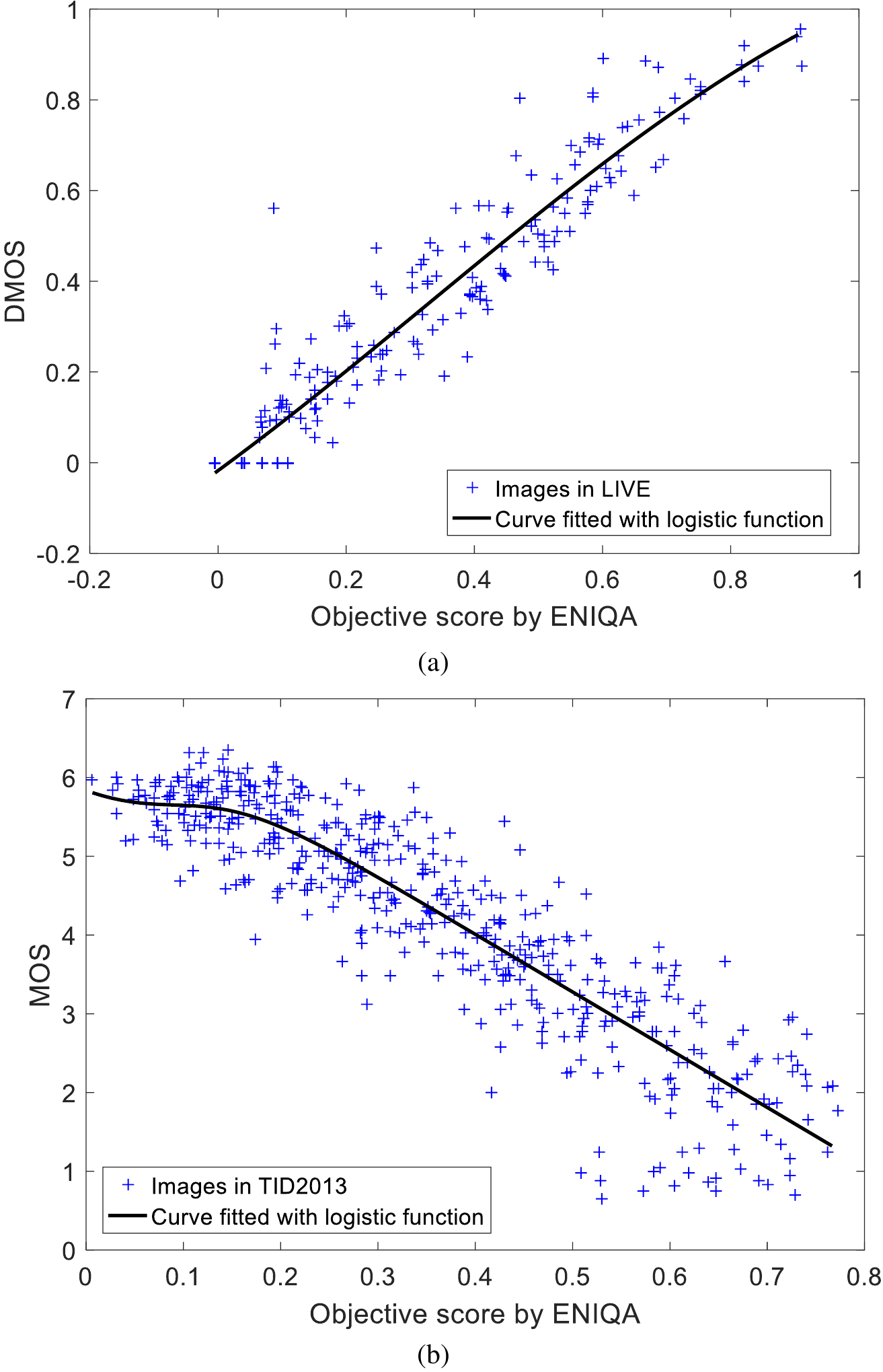}
    \caption{Scatter plots of DMOS$/$MOS versus prediction of ENIQA on the LIVE and the TID2013 databases. (a) DMOS versus prediction of ENIQA on LIVE. (b) MOS versus prediction of ENIQA on TID2013}
    \label{fig:plots}
\end{figure}

Fig.~\ref{fig:plots} shows the results of the scatter plot fitting of ENIQA on the LIVE and TID2013 databases. As in the previous experiments, when performing the scatter plot experiment on the LIVE database, we trained with the random $80\%$ of the images separated by content in the LIVE database and then tested with the remaining $20\%$, for which the results are shown in Fig.~\ref{fig:plots}(a). When conducting the experiment on the TID2013 database, we trained the model on the entire LIVE database and then tested it on the selected portion of the TID2013 database, for which the results are given in Fig.~\ref{fig:plots}(b). It can be observed from Fig.~\ref{fig:plots} that the scatter points are evenly distributed in the entire coordinate system and have a strong linear relationship with DMOS$/$MOS, which further proves the superior overall performance and generalization ability of the proposed ENIQA method.

\section{Conclusions} \label{sec:conclusion}
In this paper, we have proposed a general-purpose NR-IQA method called entropy-based no-reference image quality assessment (ENIQA). Based on the concept of image entropy, ENIQA combines log-Gabor filtering and saliency detection for feature extension and accuracy improvement. To construct an effective feature vector, ENIQA extracts the structural information of the input color images, including the MI and the TE in both the spatial and the frequency domains. The image quality score is then predicted by SVC and SVR. The proposed ENIQA method was assessed on the LIVE and TID2013 databases, and we carried out cross-validation experiments and cross-database experiments to compare it with several other FR- and NR-IQA approaches. From the experiments, ENIQA showed a superior overall performance and generalization ability when compared to the other state-of-the-art methods.


\begin{backmatter}

\section*{Acknowledgements}
The authors would like to thank Prof. Wen~Yang and Prof. Hongyan~Zhang for the valuable opinions they have offered during our heated discussions.

\section*{Funding}
This study is partially supported by National Natural Science Foundation of China (NSFC) (No. $61571334$, $61671333$), National High Technology Research and Development Program ($863$ Program) (No. $2014AA09A512$), and National Key Research and Development Program of China (No. $2018YFB0504500$).

\section*{Availability of data and material}
The MATLAB source code of ENIQA can be downloaded at~\href{https://github.com/jacob6/ENIQA}{https://github.com/jacob6/ENIQA} for public use and evaluation. You can change this program as you like and use it anywhere, but please refer to its original source.

\section*{Author's contributions}
XC conducted the experiments and drafted the manuscript. QZ and ML implemented the core method and performed the statistical analysis. GY designed the methodology. CH modified the manuscript. All authors read and approved the final manuscript.

\section*{Authors' information}
The authors are with the School of Electronic Information, Wuhan University, Wuhan 430072, China (e-mail: ygy@whu.edu.cn).

\section*{Competing interests}
The authors declare that they have no competing interests.

\end{backmatter}
\end{document}